\pdfoutput=1

\documentclass[11pt]{article}

 \usepackage{ACL2023}

\usepackage{times}
\usepackage{latexsym}

\usepackage[T1]{fontenc}

\usepackage{microtype}

\usepackage{inconsolata}

\usepackage{siunitx}
\usepackage{xcolor}
\usepackage{booktabs,colortbl, array}
\usepackage{pgfplotstable}
\pgfplotsset{compat=1.8}

\definecolor{rulecolor}{RGB}{0,71,171}
\definecolor{tableheadcolor}{gray}{0.92}
\usepackage[english]{babel}
\usepackage{amsthm}
\usepackage{amsfonts}
\usepackage{amsmath}
\usepackage{amssymb}

\usepackage{graphicx}   
\usepackage{tabularx}
\usepackage{subcaption} 
\usepackage[inline]{enumitem}
\usepackage{amsmath}

\usepackage{multirow}
\usepackage{booktabs}
\usepackage{esvect}
\usepackage{longtable}
\usepackage{makecell}

\title{Re-Representation in  Sentential Relation Extraction with Sequence Routing Algorithm}

\author{Ramazan Ali Bahrami  \and Ramin Yahyapour \\
         Georg-August-Universität Göttingen and GWDG , Göttingen, Germany\\
        \texttt{\{ramazan.bahrami, ramin.yahyapour\}@gwdg.de}}

\begin{document}
\maketitle
\begin{abstract}

Sentential relation extraction (RE) is an important task in natural language processing (NLP). In this paper we propose to do sentential RE with dynamic routing in capsules. We  first show that the proposed approach outperform state of the art on common sentential relation extraction datasets Tacred, Tacredrev, Retacred, and Conll04. We then investigate potential reasons for its good performance on the mentioned datasets, and yet low performance on another similar, yet larger sentential RE dataset, Wikidata. As such, we identify noise in Wikidata labels as one of the reasons that can hinder performance. Additionally, we show associativity of better performance with better re-representation, a term from neuroscience referred to change of representation in human brain to improve the match at comparison time. As example, in the given analogous terms King:Queen::Man:Woman, at comparison time, and as a result of re-representation, the similarity between related head terms (King,Man), and tail terms (Queen,Woman) increases. As such, our observation show that our proposed model can do re-representation better than the vanilla model compared with. To that end, beside noise in the labels of the distantly supervised RE datasets, we propose re-representation as a challenge in sentential RE 
\footnote{https://github.com/bahramiramazan/re-representation}.
\end{abstract}

\section{Introduction}

 \begin{figure}[t]
\centering
  \includegraphics[width=.85\linewidth]{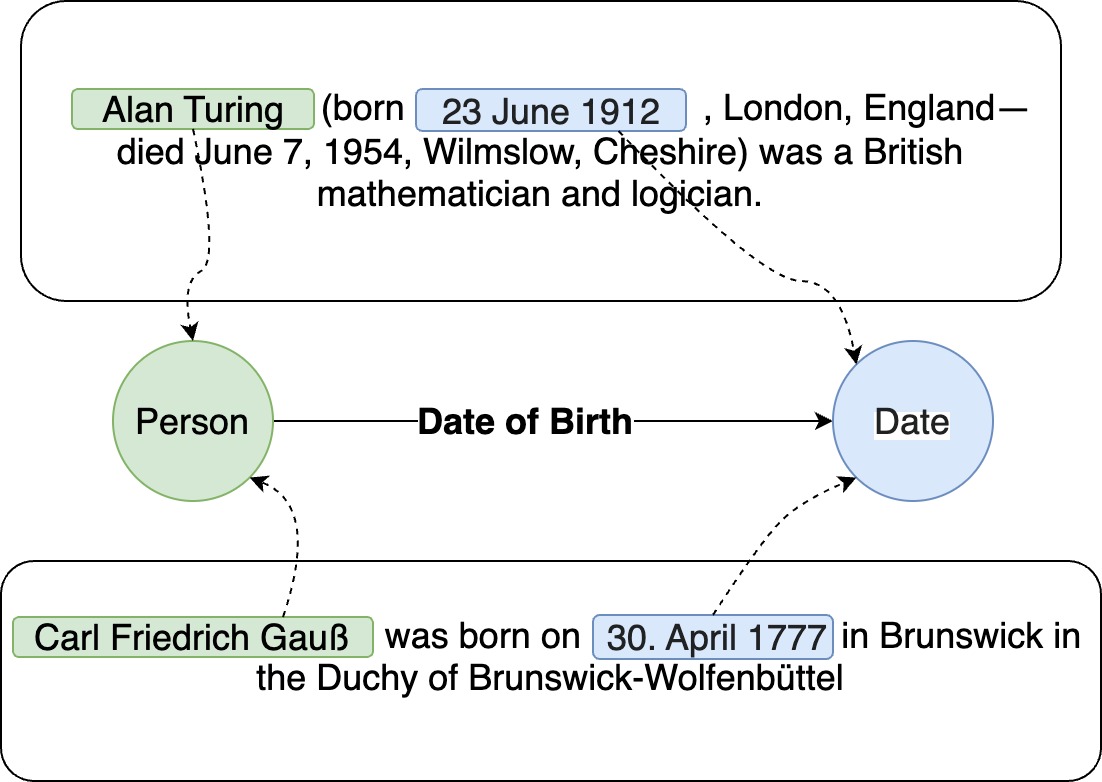} 
  \caption {The need for re-representation in sentential relation extraction (Text from Wikipedia). At comparison time, the similarity between related terms increases. Note the entity types as manual label for re-representation. Note also the proportional word analogy constructed from the given example here,  "Alan Turing": 1912 :: "C.F.Gauẞ":1777}
  \label{fig:abstraction}
\end{figure}

Sentential relation extraction is about inferring the relation between two entities in a given sentence. Various sentential relation extraction datasets are constructed with distant supervision. Accordingly, it is assumed that if two entities are related in a knowledge base such as  Wikidata, they are also related according to the sentence that contain them. As such, datasets often not only provide sentences with corresponding entities, but also additional details such as description, aliases, and types of entities. Accordingly, to improve performance, various works introduce complicated models that account for extra additional details. To that end, the results of incorporating additional details are not coherent. As example, while in some studies entity type has shown to improve performance \citep{10.1145/3442381.3449917}, in some others however, it has shown to degrade performance\citep{vashishth-etal-2018-reside}. As such, weather there are still room for improvement or if low performance are due to noise and error in the labels, is an open research question. In this paper, we aim to offer a deeper understanding of the task that can help in defining a better goal and objective for methods that incorporate the additional details into the sentential context using complex models. To do so, we propose an intuitive model that out perform state of the art on most dataset, and then investigate the reason for its better performance. To propose our approach, we build on works from neuroscience. To that end, inference such as sentential relation extraction in which a relation from one context, is mapped to a relation in another context is referred to as analogical reasoning \citep{GENTNER1983155}. In analogies, such as  proportional word analogies of the form King:Queen::Man:Woman, 'King' is related to 'Queen' as 'Man' is related to 'Woman', even though pairs (King,Man) or heads and (Queen,Woman) or tails are different. Similarly in sentential relation extraction, given that according to sentences $S_i$, for $i \in \{1,2,...,N\}$; the head entity, $e_{i}^{h}$ and tail entity, $e_{i}^{t}$ are similarly related, we have analogous terms of the form $e_{i}^{h}:e_{i}^{t}::e_{j}^{h}:e_{j}^{t}$. To that end, studies in neuroscience suggest that, comparison as the foundation of any analogical reasoning, changes the representation of objects being compared \cite{article1,Saitta2013AbstractionIA}. As example in  \citep{SILLIMAN2019128} empirical evidence of re-representation, according to which people changes the representation of entities in order to improve the match at comparison time is documented. In other words, given that analogies are about partial similarity in different contexts \citep{sun_analogy_2023}, to map a relation from one context to a relation in another context, it is therefore needed to discard some information in both contexts \citep{Saitta2013AbstractionIA} Figure \ref{fig:abstraction}. Accordingly, we found dynamic routing in  Capsules network suitable for the task.  Capsules were introduced first by \citep{10.1007/978-3-642-21735-7_6}, and dynamic routing in capsules by \citep{sabour2017dynamicroutingcapsules}. They can be thought of neurons that output different features of processed entity. We  test the proposed algorithm on relation extraction datasets  Wikidata \citep{sorokin-gurevych-2017-context}, Tacred \citep{zhang2017tacred},  Tacredrev \citep{alt-etal-2020-tacred},  Retacred \citep{stoica2021retacredaddressingshortcomingstacred}, and Conll04 \citep{roth-yih-2004-linear,eberts-ulges2019spert}.  Our observations are summarized as follows:

\begin{itemize}
\item Our proposed approach improve state of the art scores on sentential relations extraction datasets Tacred, Tacredrev, Retacred, and Conll04.
\item We estimate a significant error rate in labels of Wikidata, the dataset on which various studies try to improve model performance by incorporating the additional details through complex models.
\item We show empirical evidence of re-representation and its associativity with better sentential RE performance in neural network. 
\end{itemize}

\section{Related Works}
The use of extra additional details about entities such as entity type, aliases and description, and the way they are incorporated into the sentential context is one of the main theme of related works. As such, beside studies that addresses noise in the RE datasets, the other works deal with extra additional details and how to best incorporate them in the context. To that end, \citep{Riedel2010ModelingRA} show that the vanilla distant supervised method used for generating sentential RE datasets, result in noisy labels, and  proposes an improved version of the vanilla method, reducing error rate by 30\%. 
Additionally, in studies related to variants of the common sentential RE dataset Tacred \citep{zhang2017tacred}, Tacredrev \citep{alt-etal-2020-tacred}, and Retacred \citep{stoica2021retacredaddressingshortcomingstacred}; it is shown that after relabeling the noisy examples in Tacred, models improve performance by 8.0\% (Tacredrev) and 14.3\% (Retacred) of F1 score. Moreover, state of art performance for Tacred, and its variants, is proposed by \citep{zhou-chen-2022-improved,park2021improvingsentencelevelrelationextraction}. They show that incorporating abstract label of entities( entity types) improve model performance. Furthermore, \citep{sorokin-gurevych-2017-context} introduces Wikidata, a much larger dataset for sentential relation extraction based on the knowledge base wikidata (Table \ref{tab:statistics}). To that end, to improve performance by enriching sentential context, in addition to entity type, \citep{nadgeri-etal-2021-kgpool,10.1145/3442381.3449917} consider integrating other side information such as entity description, and aliases, through complex models such as graph neural network. Moreover, in \citep{vashishth-etal-2018-reside}, the use of entity types and relation alias information for improving performance is discussed.

\begin{figure*}[h]
  \includegraphics[width=0.98\linewidth]{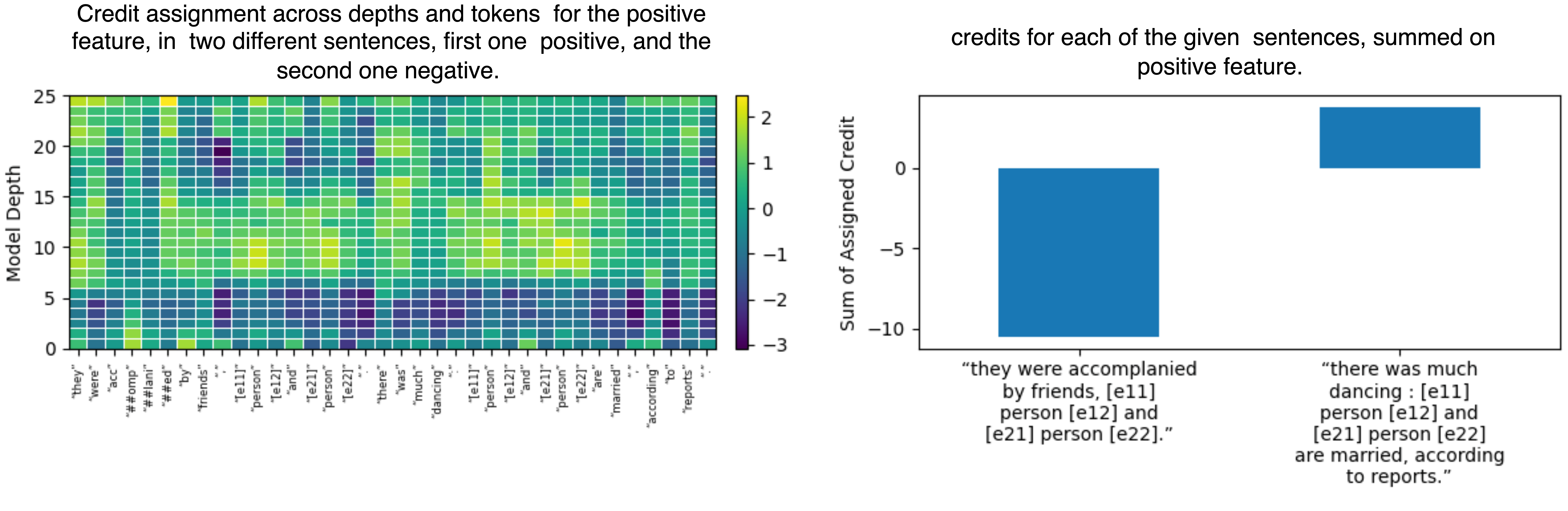} 
  \caption {Credit assignment in dynamic routing \citep{heinsen2022algorithmroutingvectorssequences}. The output here has 2 dimensions only, one for positivity, and the other for negativity. As example, given some sequence of vectors of depth h (25 here), sequence number n (number of tokens in the respective sentence), and dimension d (1024 here), and some configuration for the expected output (here depth=1, d=2, n=1), the dynamic routing algorithm works as credit assignment system. As such, projections of every feature in the input has limited credits at their disposal, and assigns it to the features in the output. Summing credits over all hidden states for positive feature, result in a value that is greater when the example is positive, and smaller otherwise(Example sentences are taken from Retacred). }
  \label{fig:credit}
\end{figure*}

\section{Problem Formulation}
\label{sec:problem-formulation}
We formulate the sentential RE based on the fundamental assumption that it is a type of analogical reasoning. To the end, in relation extraction datasets,  according to sentences $S_i$, and $S_j$  we can construct proportional word analogies of the form $e_{i}^{h}:e_{i}^{t}::e_{j}^{h}:e_{j}^{t}$, as we have  King:Queen::Man:Woman. As such, a claim based on the studies from neuroscience is that our ideal proposed model shall do re-representation\citep{SILLIMAN2019128}. More commonly, given $X^{h}_{i}:X^{t}_{i}::Y^{h}_{j}:Y^{t}_{j}$, re-representation can be viewed equivalent to a transformation $F$ such that according to some similarity measures $\psi$, when $X^{h}_{i}$ is related to $X^{t}_{i}$, as $Y^{h}_{j}$ is related to $Y^{t}_{j}$ (positive examples) , we have :

\[ 
\\
\psi (F_{}(X^{h}_{i}),F_{}(Y^{h}_{j}))\simeq 1
\]
\[ 
\psi (F_{}(X^{t}_{i}),F_{}(Y^{t}_{j}))\simeq 1
\]
and when $X^{h}_{i}$ is not related to $X^{t}_{i}$, as $Y^{h}_{j}$ is related to $Y^{t}_{j}$ (negative examples), we have:
\[ 
\\
\psi (F_{}(X^{h}_{i}),F_{}(Y^{h}_{j}))\simeq -1
\]
\[ 
\psi (F_{}(X^{t}_{i}),F_{}(Y^{t}_{j}))\simeq -1
\]

As such, the similarity function $\psi$ returns 1 when terms come from positive examples and -1 otherwise. One common example for $\psi$ is cosine similarity between two given vectors. It is to note that, as in sentential RE, the sentences containing entities express the relation between entities, it is therefore needed that any change of representation be conditioned on the contextual sentence. As such, formally the task can be presented as the minimization problem below:
\begin{flushleft}
\begin{equation*}
\min_{\substack{t \in [s,o] }} \{\psi(F(X_i^{t}\mid S_i),F(X_j^{t}\mid S_j))+(-1)^{p}
\}_{i,j=1  }^{N}
\end{equation*}
\end{flushleft}

Here N stands for the number of instances in the dataset, p=1 when both entities come from the same relation, or from positive examples, and p=0 when entities come form negative examples. Additionally, $X_i^{t}$ stands for the embeddings of an entity or word, and $S_i$ is the embeddings for the contextual sentence. Moreover, $s$ stands for subject or head entity, and $o$ for object or tail entity.

\section{Proposed Method}

\begin{figure*}[h]
\centering
  \includegraphics[width=0.98\linewidth]{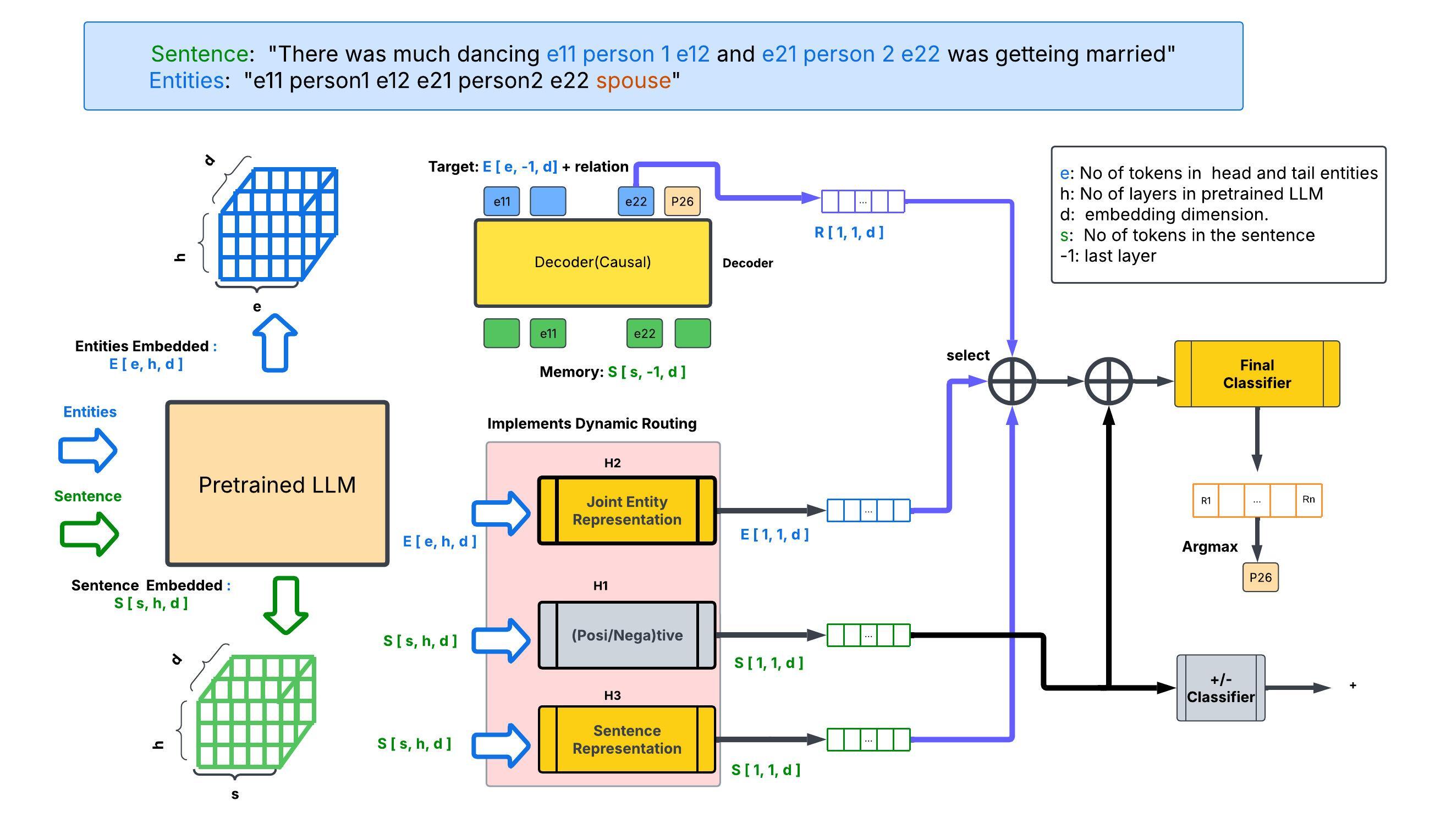} 
  \caption {Over all architecture of our proposed model.  We use different heads to classify the relation. On top is the Decoder, the head that translate from the sentence to entities post-fixed with relation id. Below it, are heads that implement routing as described in the paper \citep{heinsen2022algorithmroutingvectorssequences}. In that, H1( gray module in the middle) will identify positivity and negativity of examples. Under it is H3, the head that find the representation of the  relation or sentence with entities marked as is shown in text on top of the diagram. Above H1 is H2, the head that calculate the joint representation of concerned entities. To the left is the pre-trained large language model(LLM), the backbone from which we obtain embeddings. }
   \label{fig:architecture}
\end{figure*}

\begin{table*}[h]
    \centering

    \begin{tabular}{llllll}
   
        \toprule
        {\textbf{Dataset}}  & \thead{Tacred} & \thead{Tacredrev}   & \thead{Retacred}  & \thead{Conll04}  &\thead{wikidata}    \\
        \midrule
        No of Relations & \multicolumn{1}{c}{41 }   &  
        \multicolumn{1}{c}{41} & \multicolumn{1}{c}{40} &5 &353
        \\

        No of Abstract Entities & \multicolumn{1}{c}{23 }   &  
        \multicolumn{1}{c}{23} & \multicolumn{1}{c}{23}&4 &13533
        \\

         Train Size & \multicolumn{1}{c}{68,124}   &  
        \multicolumn{1}{c}{68,124} & \multicolumn{1}{c}{58465}&1283&372,059
        \\

        Eval Size & \multicolumn{1}{c}{22,631 }   &  
        \multicolumn{1}{c}{22,631} & \multicolumn{1}{c}{19,584}& - &-
        \\
        Test Size & \multicolumn{1}{c}{15,509 }   &  
        \multicolumn{1}{c}{15,509} & \multicolumn{1}{c}{13,418}& 422 &360,334
        \\

        Negative Size & \multicolumn{1}{c}{79.5\%  }   &  
        \multicolumn{1}{c}{79.5\% } & \multicolumn{1}{c}{79.5\% }& \textbf{-} &29\%
        \\

        \bottomrule
    \end{tabular}

        \caption{Statistics of datasets used in this work.}
    \label{tab:statistics}
\end{table*}

Our proposed model assumes an embedding model $\Omega$, and a transformation $F$ for obtaining the re-representations from the embeddings. Before giving a detailed description of our proposed method, we characterize it as follows.

\begin{itemize}

\item[1] Given some $X_i$, as tokens representing sentences, our transformation F obtains a single vector $x_{(out)}^{1d}$ of some dimension d for the joint representation of sentence containing entities $e^{h}_{i}$, and $e^{t}_{i}$, which are related to some relation $R=r_i$. 

\item[2] Instead of working with an explicit similarity function such as cosine similarity, our model is trained to maximize the following conditional probability:

\[
P_\theta\Big(R= r_{i} \mid F\big( \Omega(X_i)  \big) \Big)
\]
With $\theta$ being the model parameters.
\item[3] We show that maximizing the above conditional probability as we propose in this section will encourage explicit similarity, as was explained in Section \ref{sec:problem-formulation}.  
\end{itemize}

Given that entities and sentences are sequence of vectors, our transformation $F$ can be such, given a sequence of vectors, it outputs a single vector  $x_{(out)}^{1d}$. To that end, our proposed method for $F$ is dynamic routing in capsules. Capsules were introduced first by \citep{10.1007/978-3-642-21735-7_6}, and dynamic routing in capsules by \citep{sabour2017dynamicroutingcapsules}. In capsules network, neurons are grouped into capsules, each capsule representing some particular aspect or feature of the processed entity. Additionally, with dynamic routing, flow of data from a capsule in a layer to a capsule in the next layer depends not only on the weight matrix, but also on coefficients that itself depends on the data, also referred to as routing coefficients. The dynamic routing in capsules network is also called voting by agreement, as a capsule's vote is greater for capsules with which it agrees \citep{heinsen2022algorithmroutingvectorssequences}.

The dynamic routing algorithm used in this work \citep{heinsen2022algorithmroutingvectorssequences}, instead of voting by agreement, describe itself as credit assignment system Figure \ref{fig:credit}.

 With that being said, to obtain the embeddings of the given sequence of  words or sentence, we use some pre-trained model $\Omega$ such as bert\_base\_uncased \citep{bert} or roberta-large \citep{roberta}:
 $$
 X_{n(inp)}^{hd}=\Omega(X)
 $$
With $X$ being the tokenized sequence of words for a sentence, and $n_{(inp)}$ being the number of tokens in our input sequence, $d_{}$ being the embedding dimension, and h representing the number of hidden states in the pre-trained model.

 To generate an output $x_{(out)}^{1d}$ as representation for the given sentence, the generated multi-dimensional matrix $X_{n(inp)}^{hd}$ is feed into the sequence routing algorithm \citep{heinsen2022algorithmroutingvectorssequences}. We call the routing module as the routing head. As such, the routing head used, is configured to convert a sequence of vectors of depth h, number of sequence n, representing a sentence, term or entity into a single vector of some dimension $d$. Additionally, for experiment, we also create some of the routing heads to do some specific predefined tasks, and evaluate if adding them to the main routing head can be of help. To that end, our main routing head is used for obtaining the representation of the sentence with some marking for the concerned entities as is shown in top of the Figure \ref{fig:architecture}. Moreover, as in most datasets, a significant portions of the data are negative, we create an special routing head with two features, one representing positivity ( the relation between concerned entities is among our relation set) and another one representing negativity of examples. The working of the mentioned routing head is depicted in the Figure \ref{fig:credit}. A detailed description of routing heads and the baseline Decoder is depicted in the Figure \ref{fig:architecture}. In practice, we experiment and compare performance of a single routing head, and all routing heads combined with the Decoder, referred here as collection of experts. Each head is characterized as follows:

\begin{enumerate}
    \item H1: Obtains the representation for positivity or negativity. The output for this head has 2 dimension only, one representing positivity and another negativity Figure \ref{fig:credit}. 
    \item H2: This head learns the joint representation of head and tail entities. 
    \item H3: Is used to obtain the representation for the sentence containing concerned entities. H3 is the main routing head.  
    \item Decoder: We use a transformer based decoder for the baseline model, as is shown in the Figure \ref{fig:architecture}. As in the example in the mentioned Figure, the decoder uses the last hidden state for the sentence as memory, and entities post-fixed by the corresponding relation id as the target. 
\end{enumerate}

\subsection{Optimization}

Given the organization of our data into head and tail entities, $e^{h}_{i}$ and $e^{t}_{i}$, and  the corresponding sentences $S_{i}$ for $i \in N$, and relation $r \in R$, with R being the relation set,
and the embedding model $\Omega$, and the transformation $F$ based on dynamic routing, and an instance of data as follows:
\[
\mathcal{D} = \{(e_i^{h},e_{i}^{t},S_{i}^{}, r_{i})\}_{i=1}^{N} \quad 
\]
$$ \text{where } r_{i} \in R\quad and \quad i \in \{1, 2, 3, \ldots, N\}  $$

Where N is the dataset size. The transformation $F$  based on dynamic routing, learns a representation $x^{d}_i$, with some dimension d, such that the loss below is minimized:

\[
\mathcal{L}(\theta) = - \sum_{i=1}^{N} \log P_\theta\Big(R= r_{i} \mid     F\big( \Omega(X^{}_{i})  \big)  \Big)
\]
 Here $X^{}_{i}$ being the tokens for the sentence $S_i$, and F is the routing head. In practice, by concatenating the outputs produced by different heads, we experiment if combining heads, also referred to as collection of experts Figure \ref{fig:architecture} may be of any help. Additionally, if Decoder is among selected heads, its loss will be added to the classifiers loss as in the Figure \ref{fig:architecture}. 



\section{Experiments }

\subsection{Datasets}
We test our model on several sentential relation extraction datasets. Specifically we test the proposed model on wikidata \citep{sorokin-gurevych-2017-context}, Tacred \citep{zhang2017tacred},  Tacredrev \citep{alt-etal-2020-tacred}, Retacred \citep{stoica2021retacredaddressingshortcomingstacred}, and Conll04 \citep{roth-yih-2004-linear,eberts-ulges2019spert}. In all datasets, except Conll04, negative example make a significant portion of the examples. To that end, there are several factors to note about the datasets. 
\begin{enumerate}
    \item The ratio of positive and negative examples: Conll04 has no negative record, while wikidata has 22/29\% of example as negative, and all Tacred variants have 79.5\%  of example as negative.
    \item Number of entity types or manual abstract label of entities: All Tacred variants has 23 abstract labels for entities as and according to name entity recognition types in stanford NER system \citep{zhang2017tacred}. Conll04 has only 4 different types of entities, and Wikidata has the highest number of abstract labels for entities 13,533. 
    \item Number of relations: From number of relation points of view, wikidata has 353 relation types, which is the largest among all datasets considered, while Conll04 has only 5 types of relations. 
\end{enumerate}

\subsection{ Different Configuration of the Sentence}

\label{sec:eval}
The assumption for our proposed model is that, in order to do re-representation, dynamic routing can do feature selection . As such, we study our proposed model with different settings or configuration of the sentence. Accordingly, a given sentence can provide different level of details about the entities and their relations. As example, the sentence  "<Mask> was getting married to <Mask>.", wherein the two concerned entities are masked, provide less details as when entities are not masked. Similarly, when entities are replaced by entity type, the level of details are less than the original sentence. This is important as in order to do re-representation, models need to do abstraction, and discard some unnecessary details. As such, manual abstract label of entities or entity types can perhaps make the job of RE models on some datasets easier. To that end, the following sentence configurations are used with markings as is shown in the Figure \ref{fig:architecture}:  

\begin{itemize}
    \item Abstract: We replace surface form of the entity with entity type (abstract label of the entity). example: Germany or France is replaced by entity type Country.  
    \item Mask: We replace surface form of the entity with the placeholder, 'MASK', in the sentence.
    \item Entities: We use only surface form of the entity as is.
    \item Mix: The entity type, or abstract label for entity and its   surface form is used together with some marking. Example: "x was getting married to y." is transformed into : \textcolor{gray}{" [e11] + person * x [e12] was getting married to [e21] \# person $\&$ y \@ [e22]." } 
\end{itemize}

\begin{table}[t]
    \centering
\resizebox{0.95\linewidth}{!}{
    \begin{tabular}{llll}
   
        \toprule
        {Config} &{Model}  & \thead{Retacred } & \thead{Conll04}     \\

    \cmidrule{1-4 }
    \multirow{2}{*}{{\small Mix}}
     &H3& \multicolumn{1}{c}{ \textbf{92.2(80.1)} }   &  
        \multicolumn{1}{c}{\textbf{100.0(100.0)}}
        \\   

         &Decoder & \multicolumn{1}{c}{49.3(21.0)}   &  
        \multicolumn{1}{c}{78.6(79.8)}
        \\

   \cmidrule{2-4 }
    \multirow{2}{*}{Entities}
    
        &H3 & \multicolumn{1}{c}{\textbf{89.7(58.5)} }   &  
        \multicolumn{1}{c}{\textbf{84.1(84.7)}} 
        \\
                &Decoder& \multicolumn{1}{c}{50.4(31.5) }   &  
        \multicolumn{1}{c}{42.1(41.8)} 
        \\
     
   \cmidrule{2-4 }
    \multirow{2}{*}{{\small MASK}}
     &H3& \multicolumn{1}{c}{\textbf{ 81.7(54.2)} }   &  
        \multicolumn{1}{c}{\textbf{80.1(79.3) }} 
        \\   

         &Decoder & \multicolumn{1}{c}{--}   &  
        \multicolumn{1}{c}{--}
        \\

    \cmidrule{2-4 }
    \multirow{2}{*}{Abstract}
    
        &H3 & \multicolumn{1}{c}{\textbf{75.2(48.5)} }   &  
        \multicolumn{1}{c}{\textbf{82.2(80.3)}} 
        \\
                &Decoder& \multicolumn{1}{c}{29.1(13.0)  }   &  
        \multicolumn{1}{c}{61.8(63.7)}
        \\

        \bottomrule
    \end{tabular}
    }
        \caption{Comparative performance of the routing head H3, and transformer based Decoder on different configuration of sentence or information granularity. Recorded scores inside parenthesis are F1 Macro, and F1 Micro otherwise. The backbone model is roberta-large.}
    \label{table:granularity}

\end{table}

\subsection{Experiment One: Comparative Performance on Different Information Granularity}

We investigate performance achievable with our proposed model, and the transformer based Decoder on each sentence configuration described above. As each configuration of the sentence provide different level of details about entities and their relation, we refer to different sentence configuration as different information granularity. Accordingly, the relation between two entities in a sentence can be mostly predicted in all sentence configuration considered here; However, the best result by the proposed model is when the entity type is added to contextual sentence.
For Decoder however, the best result changes across datasets considered Table \ref{table:granularity}. As such, on Retacred, Decoder's best score is when entity type is not added to the sentence( configuration "Entities"). However on Conll04, it is the other way around. Moreover, on Retraced, Decoder have relatively low scores,  while on Conll04, our Decoder's score( 78.6 F1 Micro) is above state of the art ( with state of the art being 76.5, Tables  \ref{table:final}, and Table \ref{table:granularity}).

\subsection{Experiment Two: Entity Types as Manual Label for Re-Representation }
Given that entity types increase the similarity as is expected for re-representation (a depicted example can be seen in the Figure \ref{fig:abstraction}), we can view entity types as manual label for re-representation. To that end, we extract entities from the respective sentences, and train the proposed model on the extracted entities and entity types.  This help us study entities and entity types in isolation. In doing so, we consider all sentence configurations (except Mask) as explained in the Section \ref{sec:eval}. As such, for Conll04, our proposed model exhibit same performance  with configuration Abstract and configuration Mix Table \ref{tab:entity}. As such, we can conclude that, on some datasets,  the manual label for re-representations or entity types (configuration Abstract), result in best performance. A possible explanation would be: when the entity types or manual label of re-representation can predict the relation, or given a relation the entity types for head and tail entities can be predicted, such as in Conll04, entity types alone (config Abstract) result in peak performance and less complexity Table \ref{tab:entity}. 


\subsection{Experiment Three: Performance on Varying Number of Entity Types}

Does increasing the number of relation and entity types, or increased complexity for re-representation effect performance? To that end, we already observed the relatively good performance by Decoder on Conll04, the dataset with 4 entity types and 5 relations only Table \ref{table:granularity}. As such, we also evaluated the Decoder and the proposed model H3, on the smaller subset of Retacred, person-person, having only 6 relations and 1 entity type only. Additionally, after training on the full dataset, we recorded the performance on the same subset, person-person*. Accordingly, Decoder's performance is better when number of relation and entity types are smaller Table \ref{tab:subset}, as was noted for Conll04. As such, the experiment support the notion that transformer based Decoder changes performance across dataset presumably due to larger number of entity types, and relations. Unlike the Decoder, the proposed model exhibit relatively high performance across datasets, with different number of relation, and entity types.

\begin{table}[h]
    \centering

    \begin{tabular}{lll}
   
        \toprule
        {Dataset Subset}  & H3  & Decoder   \\
        \midrule
        Full & \multicolumn{1}{c}{92.2(80.1)}   &  
        \multicolumn{1}{c}{49.3(21.0) }
      
        \\

         Person-Person & \multicolumn{1}{c}{93.0(82.2)}   &  
        \multicolumn{1}{c}{72.6(60.4) }
      
        \\

        Person-Person* & \multicolumn{1}{c}{89.7(78.3)}   &  
        \multicolumn{1}{c}{51.6(38.2) }
      
        \\

        \bottomrule
    \end{tabular}

        \caption{Performance on varying number of relation and entity types (config mix).Values inside the parenthesis are F1 Macro, and F1 micro otherwise.  Person-Person is the subset of Retacred having head and tail entity types as person only. It is the largest subset of Retacred categorized by head-tail entity types. Full is the entire dataset. Person-Person* is  performance on the same subset, but by the model trained on the full Retacred.}
        \label{tab:subset}
\end{table}


   




\begin{table}[h]
    \centering

    \begin{tabular}{lll}
   
        \toprule
        {Metrics}  & Retacred    & Conll04    \\
        \midrule
        Mix & \multicolumn{1}{c}{\textbf{71.7}}   &  
         \multicolumn{1}{c}{\textbf{100.0} }
        \\

         Entities & \multicolumn{1}{c}{71.3}   &  
         \multicolumn{1}{c}{48.1 }
        \\

         Abstract & \multicolumn{1}{c}{62.0}   &  
         \multicolumn{1}{c}{\textbf{100.0} }
        \\

        \bottomrule
    \end{tabular}
        \caption{RE using entities extracted from the sentence, and with routing head H2. The backbone model here is roberta-large. The reported values are F1 micro. }
        \label{tab:entity}
\end{table}

\begin{figure*}[!htbp]

     \centering
     \begin{subfigure}[b]{0.99\textwidth}
         \centering
    \includegraphics[width=0.94\linewidth]{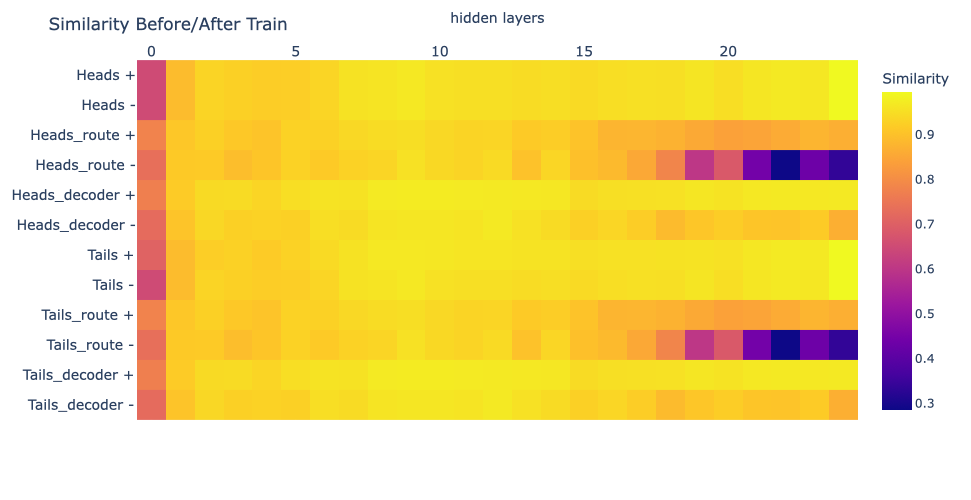} 
         \caption{As can be seen, before training, the similarity between head and tail terms in positive  (Heads +, Tails +) and negative (Heads -, Tails -) examples are barely distinguishable. However, after training, the model based on dynamic routing, does a good job of making head/tail terms more similar in positive examples, and dissimilar in negative examples. }
         \label{fig:similarity}
     \end{subfigure}

    \begin{subfigure}[b]{0.99\textwidth}
         \centering
  \includegraphics[width=0.94\linewidth]{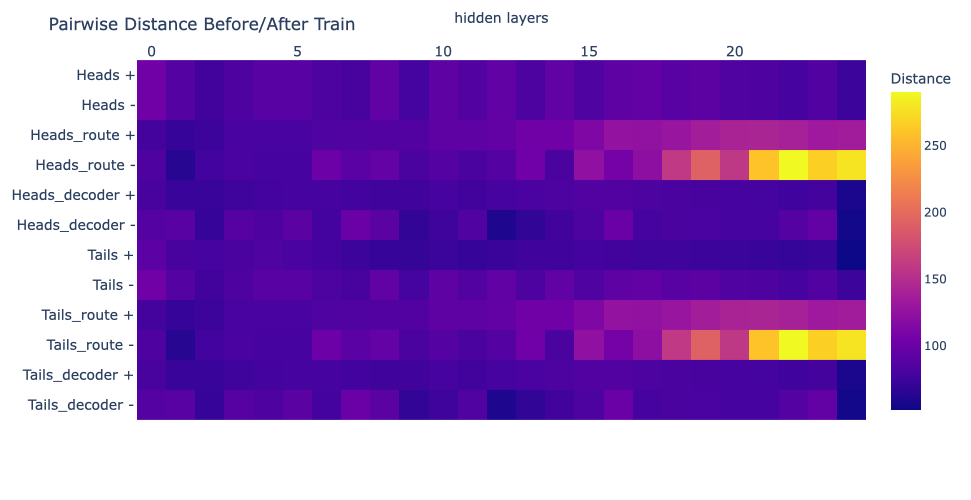} 
         \caption{The distinction between positive and negative examples are barely distinguishable before training both for head terms (Heads +, and Heads -) and also for tail terms(Tails +, Tails -). However, after training, and that also specially for the model based on dynamic routing (*\_route +, *\_route -) the increase in the distance between head/tail terms in positive examples, are far less intense than in negative examples. }
         \label{fig:semantic-distance}
     \end{subfigure}

  \caption {X-axis represent different hidden layers of the pre-trained LLM. Y-axis represent categories for which representation's similarity or distance was calculated. + represent positive analogous examples, and - represent negative analogous examples respectively. Heads and Tails are the related head and tail terms in proportional analogy. As example in king:queen::man:woman , (king, man) are head, whereas  (queen, woman) are tail. We report the result of calculations obtained on representation after training with routing heads H3(Heads/Tails\_route +/-) ,  and transformer based Decoder (Heads/Tails\_decoder -/+). We also report the same before training (Heads/Tails +/-).}

  \end{figure*}

\subsection{Experiment Four: Comparison with State of the Art}
To compare with state of the art, we trained our proposed model on the mentioned datasets, and documented the result. The result is shown in the Table \ref{table:final}. Our observations show that our proposed model outperforms state of the art on 4 datasets. To that end, our routing head H3, with roberta-large as the backbone, keeps a relatively high performance on all datasets. It outperform state of the art on all dataset, except Wikidata. In the the Section \ref{sec:wikidata} we show that noise is the main reason for the low performance on Wikidata. Moreover, despite the extra complexity that use of all heads, or expert heads, adds to our main model, we noticed little improvement. We therefore did not evaluate the expert head on Wikidata. Lastly, for our proposed model H3, the difference with bert-base-uncased \citep{bert} and roberta-large \citep{roberta} as the backbone is noticeable.

\begin{table*}[t]
    \centering

    \begin{tabular}{llllll}
   
        \toprule
        {Model}  & \thead{Tacred } & \thead{Tacredrev}   & \thead{Retacred}  & \thead{Conll04} & \thead{Wikidata}   \\
        \midrule
        Entity Marker \citeyearpar{zhou-chen-2022-improved} & \multicolumn{1}{c}{74.6 }   &  
        \multicolumn{1}{c}{83.2} & \multicolumn{1}{c}{91.1} &- &-
        \\

        Curriculum Learning\citeyearpar{park2021improvingsentencelevelrelationextraction} & \multicolumn{1}{c}{75.2 }   &  
        \multicolumn{1}{c}{-} & \multicolumn{1}{c}{\textcolor{blue}{91.4}}&- &-
        \\

       REBEL \citeyearpar{huguet-cabot-navigli-2021-rebel-relation} & \multicolumn{1}{c}{- }   &  
        \multicolumn{1}{c}{--} & \multicolumn{1}{c}{90.4}&\textcolor{blue}{76.5} &-
        \\

         KGpool \citeyearpar{nadgeri-etal-2021-kgpool} & \multicolumn{1}{c}{- }   &  
        \multicolumn{1}{c}{-} & \multicolumn{1}{c}{-}&- &\textbf{\textcolor{blue}{88.6}}
        \\

RAG4RE \citeyearpar{efeoglu2024retrievalaugmentedgenerationbasedrelationextraction} & \multicolumn{1}{c}{\textcolor{blue}{86.6 } }   &  
        \multicolumn{1}{c}{\textcolor{blue}{88.3 }  } & \multicolumn{1}{c}{73.3}&- &\textbf{\textcolor{blue}{-}}
        \\
        \midrule
        Ours bert {\tiny H3} & \multicolumn{1}{c}{84.8 (47.8)}   &  
        \multicolumn{1}{c}{85.3 (49.7)} & \multicolumn{1}{c}{89.4 (74.0)}&99.7(99.8) &84.5 (32.0)
        \\

        Ours bert {\tiny H1,H2,H3,Decoder} & \multicolumn{1}{c}{\textbf{87.4} (48.3)   } &  
        \multicolumn{1}{c}{88.7(50.9)} & \multicolumn{1}{c}{88.7 (68.5)}& \textbf{100.(100.)}  &--
        \\

    Ours Roberta  {\tiny H3}  & \multicolumn{1}{c}{87.1 \textbf{(61.1)} }   &  
        \multicolumn{1}{c}{\textbf{88.8 \textbf{(64.2)}}} & \multicolumn{1}{c}{\textbf{92.2(80.1)} }& \textbf{100. (100.)} &85.6 (32.9)
        \\   

        \bottomrule
    \end{tabular}
        \caption{Our method's performance compared with state of the art. Best score is bold, state of the art is blue. Values inside the parenthesis are F1 Macro, and F1 micro otherwise. The configuration of sentence is Mix, and backbone is as indicated. We do not test all heads(H1,H2,H3,Decoder) for Wikidata as we found H3's performance to be already good on Wikidata's noisy labels.}
    \label{table:final}
\end{table*}

\section{Observations}
\subsection{Re-Representation in Neural Network}
As suggested initially, treating sentential RE as analogy, requires some form of re-representation to improve the match. To check if neural-network also does re-representation, using a subset from Retacred test set, we create positive and negative analogous entities of the form   $e^{h}_{i}$ : $e^{t}_{i}$ :: $e^{h}_{j}$ : $e^{t}_{j}$ , for all $i,j \in \{1,2,..,N\}$ such that the corresponding sentence $S_i$ and $S_j$ expresses the same relation between the corresponding entities in positive examples and different relation in negative examples. In doing so, we obtain the embedding for a given entity in the sentence, by feeding the sentence into the backbone model, and then slice the entity from the sentence embedding. We then calculate the cosine similarity, and pairwise euclidean distance between respective head, and tail entities in both positive and  negative examples. We calculate the mentioned values across hidden states of pre-trained backbone model, both after training with each training heads H3 and  Decoder, and also before training, and then create a heat map as is shown in the Figure \ref{fig:similarity}. Accordingly, "Head +" , and "Head -" represent the cosine similarity between heads in positive and negative examples before training. As can be seen, the similarity is not much different between positive and negative examples. However, for the proposed model, after training, the similarity decreases significantly in negative examples, making the difference between positive and negative examples clearly noticeable ( specially in final layers of the backbone model). Similarly, the pairwise euclidean distance between positive and negative examples, shown in Figure \ref{fig:semantic-distance}, after training are clearly distinguishable for the proposed model(Heads\_route +, Head\_route -) than it is for the vanilla Decoder (Heads\_decoder +, Head\_decoder -).

\subsection{Noise in Wikidata's Labels}
\label{sec:wikidata}
The tow Tacred variants (Tacredrev, Retacred) are very good attempts to improve data quality and reduce error rate in the Tacred. Each of these datasets improve model performance with 8.0\% and 14.3\% F1 score over the original Tacred respectively. In comparison to Tacred, wikidata has much larger and diverse types of relations. Its quality however has not gone a similar study. Instead, a significant attention has been given in improving model performance by incorporating extra additional details through complex models. As our model's performance is below state of the art on Wikidata, we were intrigued to have a look at examples in which our model disagree with labels from the dataset. Not surprisingly though, we found out that a significant portion of errors are due to confusion in the dataset labels. As example, for instances which our model disagree with the dataset label, the labels seem random. More such examples, and statistics in Appendix \ref{Analysis:appendix-wikidata}. We categorize all examples that our model disagree with labels from dataset in the appendix \ref{Analysis:appendix-wikidata}, Table \ref{tab:categoreis} , Table \ref{tab:sample1}, and Table \ref{tab:sample2}.

\section{Limitations}
Over all the dynamic routing proposed by \citep{heinsen2022algorithmroutingvectorssequences} is efficient and scalable as is explained in the original paper.
However using all routing heads as collection of experts increases the complexity n folds, where n is the number of routing heads in the model.  However, the good news is that, perhaps a single H3 head can do a better job as is shown in the Table \ref{table:final}.

\section{Conclusion and Future Research Directions}

In this paper we improve sentential relation extraction performance on several benchmarks. Additionally we identify noise as one of the main cause for low performance on largest sentential RE benchmark Wikidata. Furthermore, we propose re-representation as one of the challenges of sentential RE models. Lastly, we show that sentential RE dataset may not be as much sentence dependent as expected \ref{Analysis:appendix-sentential}. For future research direction, we are planing to study word analogies of the form a:b::c:d, jointly with sentential RE datasets. Specifically, it would be interesting to see how much improvement can training sentential RE benchmarks bring to word analogy benchmarks. 
\section*{Ethics Statement}
We did pay for the dataset Tacred. Other datasets are publicly available. In addition for the proposed algorithm being efficient, we tried to minimize our CO2 footprint. As such, this work comply and adhere to ACL code of ethics.\footnote{\url{https://www.aclweb.org/portal/content/acl-code-ethics}} 


\bibliographystyle{acl_natbib}
\bibliography{acl2023}

\clearpage
\appendix

\section{Training and reproducibility }
\label{Training:appendix}

For all routing heads we use the code from \citep{heinsen2022algorithmroutingvectorssequences}. Additionally, we use tokenizers from https://huggingface.co for bert-base-uncased, and roberta-large respectively. Our backbone models are too from https://huggingface.co. Furthermore, for routing head H3, we train it on datasets Tacred, Retacred, and Tacredrev with batch size 64, learning rate $10^{-5}$ , and on the dataset Wikidata with batch size 128, and similar learning rate as for Tacred and its variants. For collection of experts we use an smaller batch size of 24. For optimizer with use Adam from torch.optim. Moreover, we find hidden state of routing heads to have great influence on performance. To that end, for H3, w used hidden\_d=256, and out\_dimension=512. Moreover, we trained the proposed model for Tacred and its variants for 6 epochs, while we trained only for 1 epoch on Wikidata. 

Another point to note is: In case of wikidata, when entities did not have an entity type ( or instance of as in the dataset), we checked the Wikidata knowledge base to retrieve parent class as entity type\footnote{https://query.wikidata.org} . Furthermore, when entities had several values as "instance of" or parent class, again we query Wikidata to check if they have a common parent class, and used the parent class as the entity type, if not, the most common class was uses for entity type. 
Lastly, unless explicitly mentioned, all experiments are done with Roberta-Large as backbone.

\section{Observation}
\label{Analysis:appendix}

\subsection{Are Sentential RE Datasets Truly Sentential?}
\label{Analysis:appendix-sentential}
To answer if relation between the entities, can be inferred without reading the sentence, and only be looking into entities, we trained and evaluated the proposed model on entities with configuration as was discussed for the sentence. The result for different configurations are recorded in the Table \ref{tab:entity}. Accordingly, most relation can be inferred without reading the concerned sentences.

\subsection{Noise in Wikidata's Labels}
\label{Analysis:appendix-wikidata}
On examples which our model disagree with the dataset Wikidata, we found a pattern. Specifically, given a pair (p0-p*), where p0 is label('no relation') provided by the dataset, and p*(some relation other than "no relation") predicted label, there is usually another category of predictions as (p*-p0). In both groups of examples, the probability that p * is true is similar, regardless of the label provided by the dataset. The group pairs, such as p0-p17 and p17-p0; show confusion caused as a result of incorrect labels. Some examples in Table \ref{tab:categoreis}.  

For ease of understanding, we list Wikidata relation codes used in the table with corresponding labels as: 
\begin{itemize*}
 
    \item P131(located in the administrative territorial entity) 
    \item P17(country) 
    \item P47(shares border with)
    \item P118(league)
    \item P571(inception)
    \item P47(shares border with)
    \item P361(part of )
    \item P463(member of)
\end{itemize*}

\begin{table}[h]
    \centering
    \begin{tabular}{lll}
   
        \toprule
        {\small Label-Prediction}  &  \thead{probability of\\  P* being True}  & \multicolumn{1}{c}{count }     \\
        \midrule

        P0-P17  &  
        \multicolumn{1}{c}{80.6 }
        & \multicolumn{1}{c}{4090 }
        \\

        P0-P131&  
        \multicolumn{1}{c}{90.0 }
        & \multicolumn{1}{c}{4037 }
        \\

        P0-P47   &  
        \multicolumn{1}{c}{60.0 }
        & \multicolumn{1}{c}{3518 }
        \\
        
        P0-P118  &  
        \multicolumn{1}{c}{70.0}
        & \multicolumn{1}{c}{2155 }
        \\
        
        P0-P571 &  
        \multicolumn{1}{c}{50.0}
        & \multicolumn{1}{c}{1718 }
        \\

        P0-All   &  
        \multicolumn{1}{c}{70.0}
        & \multicolumn{1}{c}{29021}
        \\

\midrule
P47-P0    &  
        \multicolumn{1}{c}{60.0 }
        & \multicolumn{1}{c}{12184}
        \\
P131-P0  &  
        \multicolumn{1}{c}{80.0}
        & \multicolumn{1}{c}{4775}
        \\     
P17-P0    &  
        \multicolumn{1}{c}{70.0}
        & \multicolumn{1}{c}{4312 }
        \\  
P361-P0   &  
        \multicolumn{1}{c}{60.0 }
        & \multicolumn{1}{c}{2152 }
        \\ 
P463-P0&  
        \multicolumn{1}{c}{70.0 }
        & \multicolumn{1}{c}{1546 }
        \\

All-P0    &  
        \multicolumn{1}{c}{60.0 }
        & \multicolumn{1}{c}{40155 }
        \\

\midrule
        label!=prediction   &  
        \multicolumn{1}{c}{- }
        & \multicolumn{1}{c}{106534}
        \\

        \bottomrule
    \end{tabular}
    \caption{Top categories(sorted) on which model's predictions does not match with the label from benchmark. * represent a relation other than 'no relation'. The probability here is calculated by sampling 10 random example from  each category, and then manually checking if p* holds.}
    \label{tab:categoreis}

\end{table}

\begin{table*}[h]
    \centering
    \begin{tabular}{ll}
   
        \toprule
        {Label-Prediction}  & Sample      \\
        \midrule
\begin{minipage}[t]{0.3\textwidth}
        \small
        P0-P17\\
        P0:No relation\\ 
        P17:country
        
\end{minipage}        
        & 
        \begin{minipage}[t]{0.3\textwidth}
        \small
       Los Dominicos is a metro station on Line 1 of the  \textcolor{blue}{Santiago Metro}  in Santiago ,  \textcolor{orange}{Chile}, and is also the eastern terminal of this line .
        \end{minipage}

        \\

\midrule

\begin{minipage}[t]{0.3\textwidth}
        \small
         P0-P131\\
        P0:No relation\\ 
        P131: Located in the \\ administrative territorial entity
        
\end{minipage}

        & 
        \begin{minipage}[t]{0.3\textwidth}
        \small
        \textcolor{blue}{Boechout} is a railway station in Boechout , \textcolor{orange}{Antwerp} , Belgium .
        \end{minipage}

        \\

    \midrule

\begin{minipage}[t]{0.3\textwidth}
        \small
         P0-P47\\
        P0:No relation\\ 
        P47:Shares border with
        
\end{minipage}

        & 
        \begin{minipage}[t]{0.3\textwidth}
        \small
There are now approximately twenty restaurants in operation in Georgia , and about nine more in North Carolina ,  \textcolor{blue}{South Carolina}  , Florida , and  \textcolor{orange}{Tennessee}.
        \end{minipage}

        \\
        
        \bottomrule
    \end{tabular}
    \caption{Random sample from $P0-P^{*}$, where $p^{*}$ is any relation from relation set other than no relation, and p0 is no relation}
     \label{tab:sample1}
\end{table*}

\begin{table*}[h]
    \centering
    \begin{tabular}{ll}
   
        \toprule
        {Label-Prediction}  & Sample    \\
        \midrule

        \begin{minipage}[t]{0.3\textwidth}
        \small
        P6-P138\\
        P6:head of government\\
        P138:named after
        \end{minipage}

        & 
        \begin{minipage}[t]{0.3\textwidth}
        \small
       She later served in the  \textcolor{blue}{Blair ministry}   under Prime Minister \textcolor{orange}{Tony Blair} in a number of roles , becoming Britains first female Foreign Secretary in 2006 .
        \end{minipage}
 \\

\midrule
        \begin{minipage}[t]{0.3\textwidth}
        \small
        P264-P136\\
        P264:record label\\
        P136:Genre
        \end{minipage}

        & 
        \begin{minipage}[t]{0.3\textwidth}
        \small
 \textcolor{blue}{CD1} is the unofficial name of an untitled album by English  \textcolor{orange}{industrial music } band Throbbing Gristle , released in 1986 through record label Mute .
        \end{minipage}
        \\
\midrule

        \begin{minipage}[t]{0.3\textwidth}
        \small
        P1416-P102\\
        P1416:affiliation\\
        P102:member of political party
        \end{minipage}

        & 
        \begin{minipage}[t]{0.3\textwidth}
        \small
Other famous Solidarity activists such as [e11] \textcolor{blue}{Anna Walentynowicz}  \textcolor{orange}{Solidarity } activists such as Anna Walentynowicz , Zbigniew Romaszewski and Antoni Macierewicz have visited the Basilica as well .
        \end{minipage}

        \\
        
        \bottomrule
    \end{tabular}
    \caption{Random sample from $p^{*}-p^{*}$, where $p^{*}$ is any relation from relation set other than no relation. Consider the first row in which both label and prediction is correct.}
    \label{tab:sample2}
\end{table*}

\end{document}